%% file: main.tex
\documentclass[letterpaper,10pt,conference]{ieeeconf}  

\IEEEoverridecommandlockouts                              

\usepackage{mymacros}
\usepackage{times} 
\usepackage{graphicx}

\usepackage[T1]{fontenc}
\usepackage{color}
\usepackage{hyperref}
\usepackage{pifont}
\usepackage{caption}

\let\labelindent\relax
\usepackage{enumitem}

\title{\LARGE \bf
Case Relation Transformer:\\ A Crossmodal Language Generation Model for Fetching Instructions
}

\author{Motonari Kambara$^{1}$ and Komei Sugiura$^{1}$
\thanks{$^{1}$The authors are with Keio University, 3-14-1 Hiyoshi, Kohoku, Yokohama, Kanagawa 223-8522, Japan.
         {\tt\small motonari.k714@keio.jp, komei.sugiura@keio.jp}
}
}

\begin{document}

\maketitle
\thispagestyle{empty}
\pagestyle{empty}

\input{abstract}
\input{section1}
\input{section2}
\input{section3}
\input{section4}

\input{section6}

\input{section7}


\bibliographystyle{IEEEtran}
\bibliography{reference}

\end{document}

%% file: abstract.tex
\begin{abstract}
There have been many studies in robotics to improve the communication skills of domestic service robots. Most studies, however, have not fully benefited from recent advances in deep neural networks because the training datasets are not large enough. In this paper, our aim is to augment the datasets based on a crossmodal language generation model. We propose the Case Relation Transformer (CRT), which generates a fetching instruction sentence from an image, such as ``Move the blue flip-flop to the lower left box.'' Unlike existing methods, the CRT uses the Transformer to integrate the visual features and geometry features of objects in the image. The CRT can handle the objects because of the Case Relation Block. We conducted comparison experiments and a human evaluation. The experimental results show the CRT outperforms baseline methods. 

\end{abstract}

%% file: section1.tex
\vspace{-1mm}
\section{Introduction
\label{sec:intro}
}
\vspace{-1mm}

Domestic service robots (DSRs) that naturally communicate with users to support household tasks are a promising solution for older adult or disabled people. There have been many studies in robotics to improve such communication skills, however most of them have not fully benefited from recent advances in deep neural networks because the corpora are not large enough. This is mainly because it is very costly to build a crossmodal corpus in which text data are grounded to real-world data such as images. In particular, text annotation of images is often done by human labelers, which is tedious and expensive. 

Given this background, in this study, we focus on augmenting text data from images. We assume that a limited number of images have been annotated by human labelers. Our aim is to augment the data by generating sentences for a given image, such as ``Grab the blue and white tube under the Coke can and move it to the right bottom box.'' Hereafter, we call this target task the fetching instruction generation (FIG) task. Such data augmentation should contribute to improving the accuracy of the crossmodal language understanding models. Indeed, Zhao et al. reported that crossmodal language generation could improve crossmodal language understanding\cite{zhao2021evaluation}.

The FIG task is challenging because ambiguity in a sentence depends not only on the target object but also on surrounding objects. Indeed, the quality of generated sentences for the FIG tasks was far worse than that for simple image captioning tasks\cite{ogura2020alleviating}. In fact, there is a big gap in the quality between reference sentences and generated sentences by conventional methods, as shown in Section~\ref{sec:exp}.

Although the FIG task was handled in previous studies\cite{ogura2020alleviating}, existing models did not handle the target object and destination simultaneously\cite{ogura2020alleviating}. Therefore, they could not generate instruction sentences, such as ``Move the blue flip flop to the lower left box.,'' where the target object is ``the blue flip flop,'' and the destination is ``the lower left box.'' Moreover, they needed a long time for training because their methods are based on the Long Short Term Memory (LSTM). In contrast, in this paper, we propose a model based on the Transformer encoder--decoder\cite{vaswani2017attention}.

In this paper, we propose the Case Relation Transformer (CRT), which generates a fetching instruction sentence that includes a spatial referring expression of a target object and a destination. The CRT consists of three main modules: Case Relation Block (CRB), Transformer encoder, and Transformer decoder. Unlike existing methods\cite{ogura2020alleviating}, the CRT uses the Transformer to integrate the visual and geometry features of objects appearing in the FIG task. Moreover, unlike the Object Relation Transformer (ORT\cite{herdade2019image}), the CRT can handle both the target object and destination.

\begin{figure}[t]
    \centering
    \includegraphics[width=\linewidth]{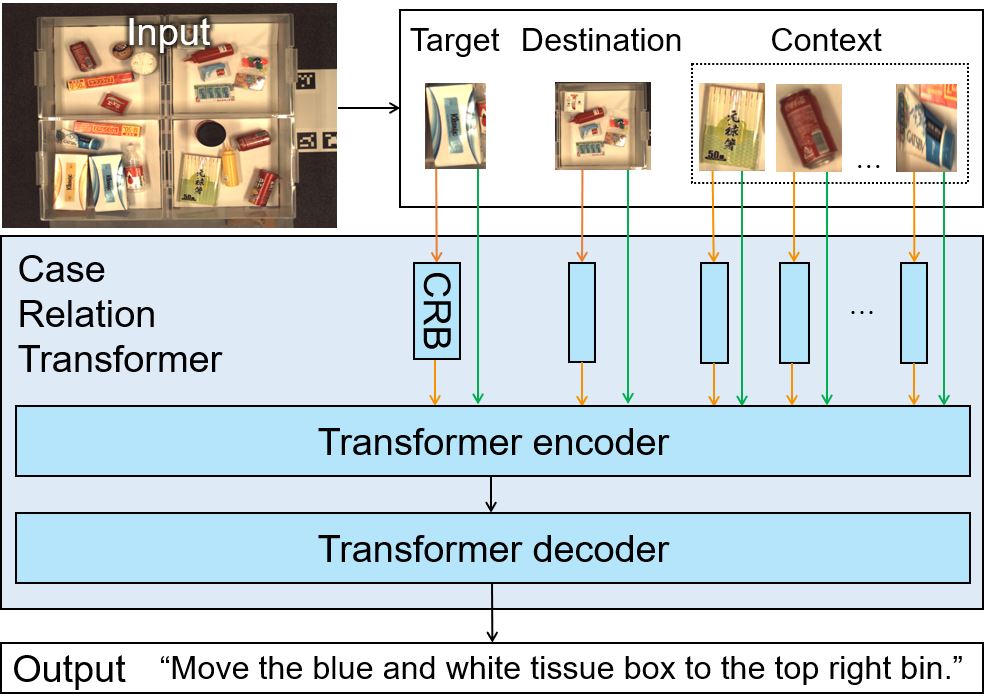}
    \vspace{-5mm}
    \caption{Overview of the CRT: the CRT generates fetching instructions from given input images. CRB represents the Case Relation Block.}
    \label{fig:eye_catch}
    \vspace{-3mm}
\end{figure}

Fig.~\ref{fig:eye_catch} shows a schematic diagram of the approach. In the figure, CRB, orange arrows, and green arrows represent the CRB,  the visual features, and the geometry features, respectively. A demonstration video is available at this URL\footnote{\protect\href{https://youtu.be/Tmzxmz2BeBg}{https://youtu.be/Tmzxmz2BeBg}}.

The main contributions of this paper are as follows:
\begin{itemize}
 \item We propose a crossmodal language generation model, the CRT, for the FIG task based on the Transformer model. 
 \item We extend the ORT by introducing the CRB to handle the target object, destination, and context information.
\end{itemize}

%% file: section2.tex
\vspace{-1.4mm}
\section{Related Work
\label{related}
}
\vspace{-0.8mm}
There have been many attempts on crossmodal language processing. Baltrušaitis et al. surveyed recent advances in crossmodal language processing\cite{baltruvsaitis2018multimodal}. The paper summarizes representative models (e.g., \cite{kiros2014unifying}\cite{mao2014deep}). Mogadala et al. also provided a comprehensive summary of tasks, existing research, and standard datasets in the vision and language area\cite{mogadala2019trends}. 

As explained in Section~\ref{sec:intro}, we focus on the fetching instruction generation (FIG) task. The FIG task is one of the vision and language tasks, and is very relevant to the image captioning task\cite{ogura2020alleviating}.

The image captioning task is closely related to the conditional language model\cite{de2015survey} and natural language generation\cite{reiter2000building}. The goal of the image captioning task is to give a human-readable description to the input image\cite{mogadala2019trends}. In the image captioning task, there are some methods that  combine attention mechanism and reinforcement learning (e.g., \cite{zhao2018multi}\cite{gu2018stack}).

There are some models regarding the FIG task in the WRS-PI dataset\cite{ogura2020alleviating}, such as the Multi-ABN\cite{magassouba2020multimodala} and the ABEN\cite{ogura2020alleviating}. The Multi-ABN raised the issue of automating dataset generation for DSRs. The MTCM-AB\cite{magassouba2020multimodalb} is a crossmodal language understanding model that identifies the target object from the instruction sentences. It dealt with the task that connects the crossmodal language processing and the DSRs. 

There are several standard datasets used in vision and language tasks\cite{mogadala2019trends}. These datasets contain images and captions, which consist of descriptive text about the images. These datasets are mainly used for image captioning tasks. In contrast, our aim is to generate instruction sentences for the target object in the image. Therefore, in this study, we use a dataset that explicitly includes a target object. 

There are some crossmodal language processing models using the Transformer\cite{vaswani2017attention} (e.g., \cite{lu2019vilbert}\cite{chen2020uniter}). The VLN-BERT\cite{majumdar2020improving} is a crossmodal language understanding transformer-based model for scoring the compatibility between an instruction and a sequence of panoramic RGB images captured by the agent.

The VL-BERT\cite{su2019vl} is similar to the CRT in terms of a crossmodal language processing model that utilizes geometry features. Unlike CRT, the VL-BERT cannot generate fetching instruction sentences which include the destination. The ORT\cite{herdade2019image} is also a crossmodal language generation model using Transformer, which uses geometry features in the same way as the VL-BERT. The ORT is similar to the CRT in that both of them handle positional relationships of objects, however, the ORT cannot handle the target object and destination. This means the ORT could not generate an instruction sentence, such as ``Move the blue flip-flop to the lower left box.''  

%
%



%

%% file: section3.tex
\vspace{-1.0mm}
\section{Problem Statement
\label{sec:problem}
}
\vspace{-0.8mm}

\vspace{-0.5mm}
\subsection{Task description}
\vspace{-1mm}

This study targets crossmodal language instruction generation for DSRs, that is, the FIG task. We assume that the sentences contain spatial referring expressions. The spatial referring expressions are referring expressions described by the positional relationship between the object being described and the objects around it to identify the object. Specific examples of them include ``books on the desk'' and ``cans next to the plastic bottles.'' In this task, the desired behavior is to generate an unambiguous instruction sentence that can accurately identify the target object and the destination for a given image.

\begin{figure}[tb]
    \centering
    \includegraphics[clip, width=60mm]{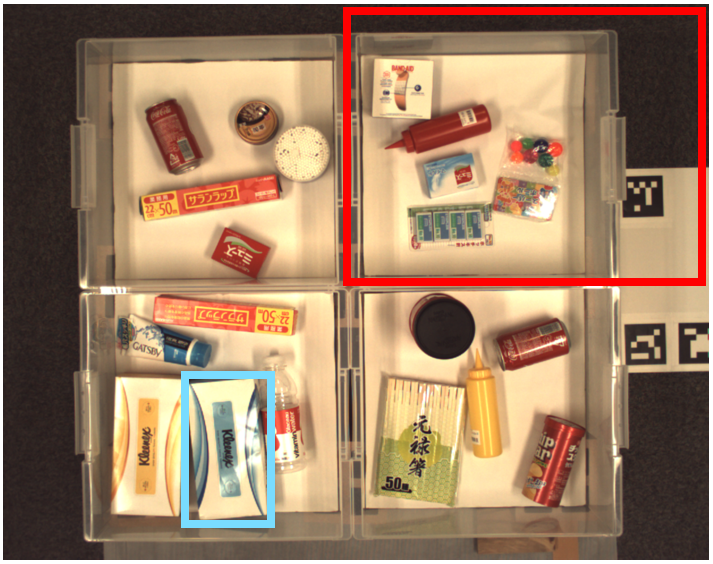}
    \vspace{-1mm}
    \caption{Typical scene of the FIG task. The target object is the blue and white tissue box (light blue) and the destination is top right (red). An instruction sentence like ``Move the blue and white tissue box to the top right bin.''  should be generated.}
    \label{fig:pfn-pic}
    \vspace{-3mm}
\end{figure}

Fig.~\ref{fig:pfn-pic} shows a typical scene of the FIG task. In this figure, light blue and red bounding boxes represent the target object and the destination, respectively. The target object is the blue and white tissue box and the destination is the top right corner of the view. In this case, an instruction sentence like ``Move the blue and white tissue box to the top right bin.''  should be generated.

The task is characterized by the following:
\begin{itemize}
    \item \textbf{Input:} 
        \begin{itemize}
            \item An image containing a target object, a destination, and context information.
            \item The coordinates of the regions of the target object and destination.
        \end{itemize}
    \item \textbf{Output:} 
    \begin{itemize}
        \item An instruction sentence to move the target object to the destination.
    \end{itemize}
\end{itemize}
The inputs are explained in detail in Section~\ref{method}.

We define the terms used in this paper as follows:
\begin{itemize}
    \item \textbf{Target object:}  an everyday object, e.g., a bottle or can, that is to be grasped by the robot.
    \item \textbf{Destination:} one of the four moving directions for the target object: top right, bottom right, top left, or bottom left. 
    \item \textbf{Context information:}  the set of regions in the image detected by an object detector (e.g., Up-Down Attention\cite{anderson2018bottom}).
\end{itemize}

\input{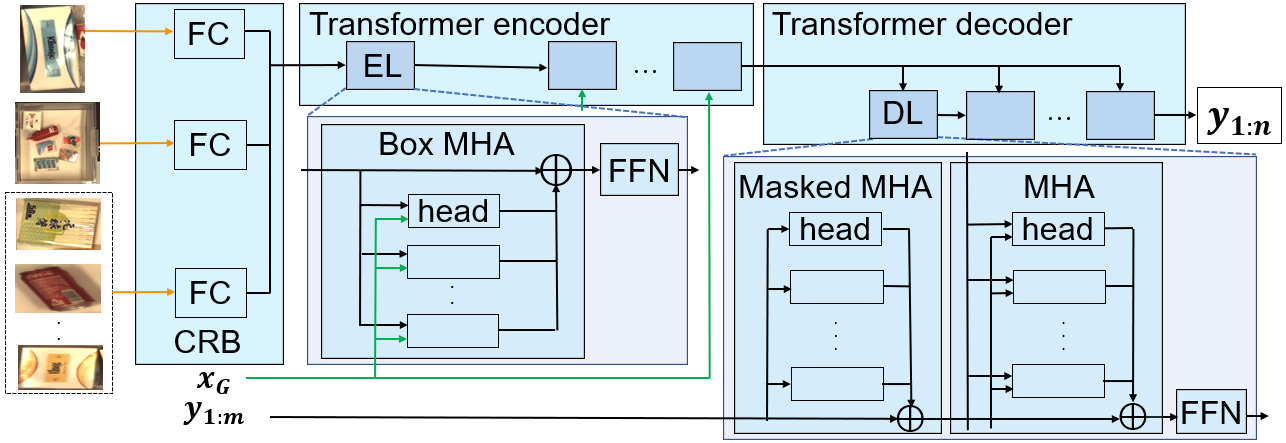}
We use the Up-Down Attention model pre-trained on the MSCOCO dataset\cite{lin2014microsoft} to extract the context information. We assume that up to 30 objects are detected in a scene image.
We evaluate the generated sentences using standard automatic evaluation metrics for natural language generation tasks such as BLEU4\cite{papineni2002bleu}, ROUGE-L\cite{lin2004rouge}, METEOR\cite{banerjee2005meteor},  CIDEr-D\cite{vedantam2015cider}, and SPICE\cite{anderson2016spice}. 
These metrics are commonly used for image captioning studies.  
We also use the Mean Opinion Score (MOS) for human evaluation.


%% file: fig/network.tex
\begin{figure*}[tb]
    \centering
    \includegraphics[clip,height=50mm]{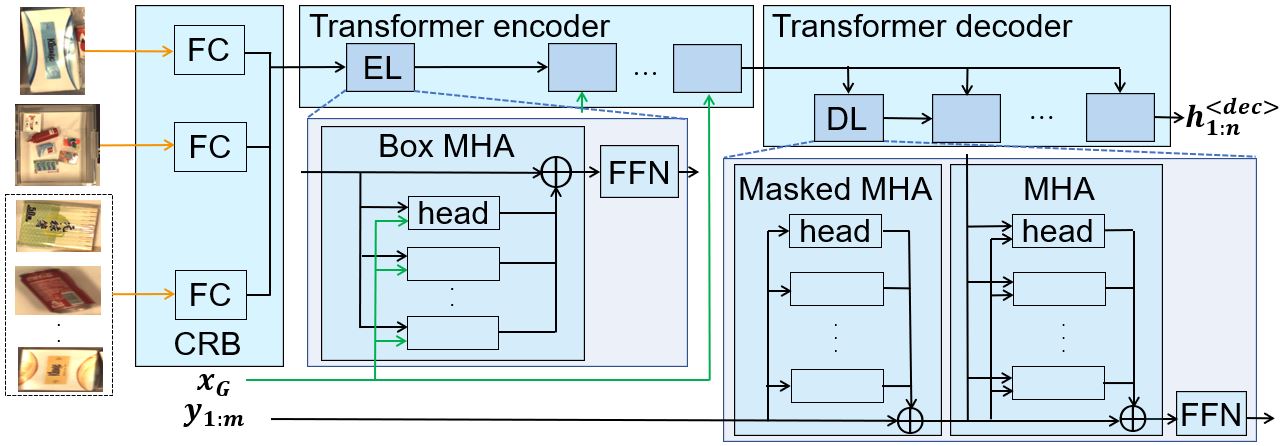}
    \vspace{-1.5mm}
    \caption{The network structure of the CRT. The CRT comprises the Case Relation Block (CRB), Transformer encoder, and Transformer decoder. FC, EL, FFN, DL, MHA, and ``head'' represent the the fully connected layer, the encoder layer, the feedforward network layer, the decoder layer, the multi-head attention layer, and the attention head\cite{vaswani2017attention}, respectively.}
    \label{fig:encoder-decoder}
    \vspace{-4mm}
\end{figure*}

%% file: section4.tex
\vspace{-1mm}
\section{Proposed Method
\label{method}
}

\vspace{-0.8mm}
\subsection{Novelty}
\vspace{-1mm}
Our model is based on the Object Relation Transformer (ORT\cite{herdade2019image}) because it can model spatial referring expressions based on the relative positions of objects. We use the Up-Down Attention\cite{anderson2018bottom} for feature extraction according to \cite{herdade2019image}. The attention mechanism enables spatial referring expressions to be acquired.
The CRT differs from existing methods in the following ways:
\begin{itemize}
    \item Unlike the ABEN\cite{ogura2020alleviating}, which follows an encoder--decoder structure based on the ResNet-50\cite{he2016deep} and the LSTM, the CRT is based on the Transformer encoder--decoder\cite{vaswani2017attention}.
    \item The difference between the ORT and the CRT is whether a target object and destination can be handled. In the CRT, the Case Relation Block (CRB) is introduced to accommodate those information.
\end{itemize}

\vspace{-2mm}
\subsection{Input}
\vspace{-1mm}
The network inputs $\bm{x}$ are defined as follows:
\vspace{-0.5mm}
\begin{align}
    \bm{x} &= (\bm{x}^{<targ>},\  \bm{x}^{<dest>},\  \bm{X}^{<cont>}), \nonumber \\
    \bm{X}^{<cont>} &= (\bm{x}^{<1>},\  \bm{x}^{<2>}, ... , \bm{x}^{<N>}), \nonumber \\
    \quad \bm{x}^{<i>} &= (\bm{x}_{V}^{<i>}, \bm{x}_{G}^{<i>}), 
\vspace{-0.5mm}
\end{align}
where $\bm{x}^{<targ>},\  \bm{x}^{<dest>},\  \bm{X}^{<cont>}$, and $N$ denote the features of a target object, destination, context information, and the number of objects contained in the context information, respectively. $\bm{x}_{V}^{<i>}$ and $\bm{x}_{G}^{<i>}$ denote visual features and geometry features of the $i$-th object, respectively.

We use ResNet-50 for preprocessing the target object and destination information. The output from the conv5\_x layer is used as $\bm{x}_{V}^{<targ>}$ and $\bm{x}_V^{<dest>}$. We extracted $1 \times 2048$ dimensional features for each of the target object, destination, and objects contained in the context information.

When preprocessing the context information, we use the same procedure for object detection and feature extraction as \cite{anderson2018bottom}. We use the Faster R-CNN\cite{ren2016faster} for object detection, and we use the output from the conv5\_x layer in the ResNet-101 as $\bm{X}_{V}^{<cont>}$.

The geometry feature of the $i$-th object, $\bm{x}_{G}^{<i>}$, is defined as follows: 
\vspace{-0.5mm}
\begin{align}
    \bm{x}_{G}^{<i>} = [r^{<i>}_{xmin}/W,\ r^{<i>}_{ymin}/H,\  r^{<i>}_{xmax}/W,\  r^{<i>}_{ymax}/H], 
\vspace{-0.5mm}
\end{align}
where $(r^{<i>}_{xmin}, r^{<i>}_{ymin}, r^{<i>}_{xmax}, r^{<i>}_{ymax})$ denotes the minimum $x$ coordinate, the minimum $y$ coordinate, the maximum $x$ coordinate, and the maximum $x$ coordinate for each region. Further, $W$ and $H$ denote the width and height, respectively.

\vspace{-2mm}
\subsection{Structure}
\vspace{-1mm}

Fig.~\ref{fig:encoder-decoder} shows the network structure of the CRT. In the figure, $\bm{x}_G$, and $\bm{y}_{1:m}$ denote the geometry features and a sentence, respectively. The orange and green arrows represent the visual and geometry features, respectively, and $\bm{h}^{<dec>}_{1:n}$ represents a token sequence. In the figure, FC, EL, FFN, DL, MHA, and ``head'' represent the fully connected layer, encoder layer, feedforward network layer, decoder layer, multi-head attention layer, and attention head\cite{vaswani2017attention}, respectively. 

The CRT consists of three main modules: the CRB, Transformer encoder, and Transformer decoder. The details of each module are described below. 

\subsubsection{Case Relation Block}
The CRB first transforms the inputs $\bm{x}$ linearly using a fully connected (FC) layer, $f_{FC}(\cdot)$. The output $\bm{h}_V \in \mathbb{R}^{N \times d_{model}}$ is obtained as follows: 
\vspace{-0.5mm}
\begin{align}
    \bm{h}_V &= \left\{ f_{FC}(\bm{x}_{V}^{<targ>}), f_{FC}(\bm{x}_{V}^{<dest>}), f_{FC}(\bm{x}_{V}^{<cont>}) \right\},\nonumber
\vspace{-0.5mm}
\end{align}
where $d_{model}$ denotes a dimension (e.g., 512).
We concatenate them in this order to condition the output. $\bm{h}_G \in \mathbb{R}^{N \times 4}$ is also obtained by concatenating $\bm{x}_{G}^{<i>}$ $(i=1,2, …, N)$ in this order. $\bm{h}_V$ and $\bm{h}_G$ are input to the Transformer encoder.

\subsubsection{Transformer encoder} 


The Transformer encoder consists of six encoder layers, each of which consists of a box multi-head attention layer and the FFN layer. The inputs to each encoder layer are $\bm{h}_{in}^{<el>}$ and $\bm{h}_G$, where $\bm{h}_{in}^{<el>}$ is the output of the previous encoder layer, and $\bm{h}_V$ for the first layer. 

First, in the box multi-head attention layer, a displacement vector $\bm{\Lambda}(m,n)$ for regions $m$ and $n$ is calculated from their information as follows:
\vspace{-0.5mm}
\begin{align}
       \bm{\Lambda}(m, n) &= \{\lambda(\delta w, w_m), \lambda(\delta h, h_m), \lambda(w_n, w_m), \lambda(h_n, h_m)\}, \nonumber  \\
       \lambda(x, y) &= \log(x/y), \nonumber
\vspace{-0.5mm}
\end{align}
where $w_i$ and $h_i$ denote the width and hight of a region $i$, respectively. $\delta w$ and $\delta h$ denote $|r^{m}_{xmin}-r^{n}_{xmin}|$ and $|r^{m}_{ymin}-r^{n}_{ymin}|$, respectively. 
The geometric attention weights $\bm{\omega}_G^{mn}$ are then calculated as follows:
\vspace{-0.5mm}
\begin{align}
    \bm{\omega_G}^{mn} &= {\rm ReLU}(f_{em}(\bm{\Lambda}(m,n)\bm{W}_G), 
\vspace{-0.5mm}
\end{align}
where $f_{em}(\cdot)$ denotes the positional encoding used in the Transformer. Query $\bm{Q}_E$, Key $\bm{K}_E$, and Value $\bm{V}_E$ are also calculated as follows: 
\vspace{-0.5mm}
\begin{align}\label{eq:qkv}
    \bm{Q}_E &= \bm{W}_{qe}\bm{h}_{in}^{<el>},\nonumber \\
    \bm{K}_E &= \bm{W}_{ke}\bm{h}_{in}^{<el>},\\
    \bm{V}_E &= \bm{W}_{ve}\bm{h}_{in}^{<el>},\nonumber
\vspace{-0.5mm}
\end{align}
where $\bm{W}_{qe}$, $\bm{W}_{ke}$, and $\bm{W}_{ve}$ denote the weights for the $\bm{Q}_E$, $\bm{K}_E$, and $\bm{V}_E$, respectively.

Next, $\bm{\omega}_G^{mn}$, $\bm{Q}_E$, $\bm{K}_E$, and $\bm{V}_E$ are input to $N_E$ attention heads. At this time, $\bm{Q}_E$, $\bm{K}_E$, and $\bm{V}_E$ are divided into $N_E$ pieces. The attention heads are implemented in parallel inside the box multi-head attention layer. In each attention head, the visual-based attention weights $\bm{\omega}_A^{mn}$ is calculated as follows:
\vspace{-0.5mm}
\begin{align} \label{eq:weight}
    \bm{\omega}_A^{mn} &= \frac{\bm{Q}_E\bm{K}_E^\top}{\sqrt{d_k}}, 
\vspace{-0.5mm}
\end{align}
where $d_k = d_{model}/N_E$ denotes the number of dimensions of $\bm{K}_E$. 
the attention weights $\bm{\omega}^{mn} \in \mathbb{R}^{N \times N}$ are calculated from the visual-based attention weights $\bm{\omega}_A^{mn}$ and $\bm{\omega}_G^{mn}$ by the softmax function as follows: 
\vspace{-0.5mm}
\begin{align}\label{eq:saweight}
     \bm{\omega}^{mn} &= \frac{\bm{\omega}_G^{mn}\exp{\bm{ \omega}_A^{mn}}}{\sum_{l=1}^{N}\bm{\omega}_G^{ml}\exp{\bm{ \omega}_A^{ml}}}, 
\vspace{-0.5mm}
\end{align}
where $\bm{\omega}_A^{mn}$ is calculated from $\bm{h}_{in}^{<el>}$ using \eqref{eq:weight}. 
Then, the self-attention $\bm{h}_{sa}$, which is the output of each head is calculated as follows. 
\vspace{-0.5mm}
\begin{align} \label{eq:sa}
\bm{h}_{sa} &= f_{sa}(\bm{Q}_E, \bm{K}_E, \bm{V}_E) \nonumber \\
    &= \bm{\omega}^{mn}\bm{V_E}.
\vspace{-0.5mm}
\end{align}
The outputs from the $N_E$ attention heads are concatenated and transformed as follows:
\vspace{-0.5mm}
\begin{align}
\bm{h}_{mh} &= f_{mh}(\bm{Q}_E, \bm{K}_E, \bm{V}_E) \nonumber \\
    &= \{\bm{h}_{sa}^{<1>}, \bm{h}_{sa}^{<2>}, ...,\bm{h}_{sa}^{<N_E>}\}\bm{W}^M, \label{eq:mh} \\
\bm{h}_{sa}^{<i>} &= f_{sa}(\bm{Q}_E\bm{W}^{Q}_i, \bm{K}_E\bm{W}^{K}_i, \bm{V}_E\bm{W}^{V}_i),  \nonumber 
\vspace{-0.5mm}
\end{align}
where $\bm{W}^M$ denotes the weight for the $\bm{h}_{mh}$.
$\bm{h}_{mh}$ is combined with the inputs $\bm{h}_{in}^{<el>}$. The output of each encoder layer $\bm{h}_{out}^{<el>}$ is calculated as follows:
\vspace{-0.5mm}
\begin{align}\label{eq:ffn}
        \bm{h}_{out}^{<el>} &= f_{FFN}(\bm{h}_{mh}), 
\vspace{-0.5mm}
\end{align}
where $f_{FFN}(\cdot)$ denotes the FFN layer. $\bm{h}_{out}^{<el>}$ is then input to the Transformer decoder.

\subsubsection{Transformer decoder}
The decoder is a stack of six decoder layers consisting of a masked multi-head attention layer, a multi-head attention layer, and the FFN layer. The inputs of the Transformer decoder are $\bm{h}^{<dec>}_{in}$ and $\bm{y}_{1:m}$, which are the output from the encoder $\bm{h}_{out}^{<enc>}$ and the sentence, respectively. $\bm{y}_{1:k}$ and $\bm{h}^{<dec>}_{1:k}$ $(k \in \mathbb{N})$ represent the following sequence: 
\vspace{-0.5mm}
\begin{align}
        \bm{y}_{1:k} &= (\bm{y}_1, \bm{y}_2, ..., \bm{y}_k), \nonumber \\
        \bm{h}^{<dec>}_{1:k} &= (\bm{h}^{<dec>}_1, \bm{h}^{<dec>}_2, ..., \bm{h}^{<dec>}_k).
\vspace{-0.5mm}
\end{align}
In training, we set $m = N$, where $N$ denotes a sentence length. In testing, we set $m = 1:j-1$, $n=j$ to predict the $j$-th word. 

First, in the masked multi-head attention layer, $\bm{y}_{1:m}$ is masked to prevent the decoder from receiving the ground truth. The self-attention of $\bm{y}_{1:m}$, $\bm{h}_{emb}$, is calculated in the same way as from \eqref{eq:qkv} to \eqref{eq:sa}, where $\bm{h}_{in}^{<el>}$ and $\bm{\omega}^{mn}$ are replaced by $\bm{y}_{1:m}$ and ${\rm softmax}(\bm{\omega}_A^{mn})$. Next, in the multi-head attention layer, $\bm{Q}_D$, $\bm{K}_D$, and $\bm{V}_D$ are calculated as follows: 
\vspace{-0.5mm}
\begin{align}
        \bm{Q}_D &= \bm{W}_{qd}\bm{h}_{emb}, \nonumber \\
        \bm{K}_D &= \bm{W}_{kd}\bm{h}^{<dec>}_{in}, \\
        \bm{V}_D &= \bm{W}_{vd}\bm{h}^{<dec>}_{in}, \nonumber
\vspace{-0.5mm}
\end{align}
where $\bm{W}_{qd}$, $\bm{W}_{kd}$, and $\bm{W}_{vd}$ denote the weights for the $\bm{Q}_D$, $\bm{K}_D$, and $\bm{V}_D$, respectively.
Then the same processing as in the Equations (8) and (9) are performed. For this processing, we can generate an embedded token sequence using the information acquired by the encoder. The output is denoted the token sequence $\bm{h}^{<dec>}_{1:n}$. The token sequence $\bm{h}^{<dec>}_{1:n}$, which is the output of the decoder layer, is calculated by the FFN layer.

\subsubsection{Generator}
The input to the Generator is $\bm{h}_g \in \mathbb{R}^{1 \times 512}$. $\bm{h}_g$ is set as $\bm{h}^{<dec>}_n$ in testing, and $\bm{h}^{<dec>}_{1:n}$ in training. The probability $p(\bm{\hat{y}})$ is then calculated for each word in the dictionary as follows:
\vspace{-0.5mm}
\begin{align}
    p(\bm{\hat{y}}) &= {\rm softmax}(f_{FC}(\bm{h}_g)), 
\vspace{-0.5mm}
\end{align}
where $\bm{\hat{y}}$ denotes the predicted sentence sequence. This word prediction is performed sequentially.

We use the following loss function:
\vspace{-0.5mm}
\begin{align} \label{equ:L}
    L = - \frac{1}{I}\sum_{i=1}^{I}\sum_{j=1}^{J}\log (p(\hat{y}_{ij})),
\vspace{-0.5mm}
\end{align}
where $I$ and $J$ denote the number of samples and length of each sentence, and $i$ and $j$ denote their indexes. Further, and $p(\hat{y}_{ij})$ denotes prediction probability of the predicted word $\hat{y}_{ij}$, respectively. The sample is defined as a set consisting of a target object, destination, context information, and an instruction sentence.

We set $m = N$  and $n = N$ in training, where $N$ denotes a sentence length, and set $m = j-1$ and $n = j$ in testing, where $j$ denotes the number of words to be predicted.

%% file: section6.tex
\section{Experiments
\label{sec:exp}
}
\vspace{-1mm}
\subsection{Dataset}
\vspace{-1mm}
\input{tab/params}

From the viewpoint of reproducibility, we use the standard dataset, PFN-PIC, which is publicly available\cite{hatori2018interactively}. The dataset contains information about the target object's region and the destination, which is sufficient for the evaluation in this study. In the experiments, we validated our model with the PFN-PIC dataset.

The PFN-PIC dataset contains object manipulation instructions in English\cite{hatori2018interactively}. The sentences were given by annotators via crowdsourcing using Amazon Mechanical Turk. They were asked to give sentences using intuitive and unambiguous expressions about the target object and destination. The dataset contains ambiguous or incorrect expressions, as reported in \cite{hatori2018interactively}. Such expressions were not filtered out in this study.

The dataset consists of the coordinates of the target objects, information about the destination (top right, bottom right, top left, or bottom left), and instruction sentences. In this study, we did not make any changes to the dataset itself except for small changes to the format.

The PFN-PIC dataset was annotated with instruction sentences by at least three annotators for each target object. The dataset is divided into two parts: a training set and validation set. The training set consists of 1,180 images, 25,900 target objects, and 91,590 instructions. The validation set consists of 20 images, 352 target objects, and 898 sentences of instructions.

We split the training set of the PFN-PIC dataset into a training set and validation set. We used the training set for training the parameters and the validation set for selecting the appropriate hyperparameters. We used the validation set of the PFN-PIC dataset as the test set and used it to evaluate the performance. The training, validation, and test sets contained 81,087, 8,774, and 898 samples, respectively.

\vspace{-0.5mm}
\subsection{Parameter settings}
\vspace{-1mm}

The parameter settings of the CRT are shown in Table~\ref{tab:parameters}. In this table, ``\#Layers'' represents the number of encoder layers in the Transformer encoder and decoder layers in the Transformer decoder, and ``\#Attention heads'' represents the number of attention head layers in a multi-head attention layer. The values of hyperparameters were based on the ORT. 
The total number of trainable parameters was 59 million.
The training was conducted on a machine equipped with a GeForce RTX 2080 Ti with 11 GB of GPU memory, 64 GB RAM, and an Intel Core i9 9900K processor. The feature extraction by Up-Down Attention\cite{anderson2018bottom} was performed on a machine equipped with a TITAN RTX with 24 GB of GPU memory, 256 GB RAM and an Intel Core i9 9820X processor.

Pre-training and fine-tuning took approximately one hour. To select the best model, we evaluated the performance of the validation set every 3,000 samples during training. We saved the model with the highest SPICE\cite{anderson2016spice} score and used it for the evaluation on the test set.

\vspace{-1mm}
\subsection{Quantitative results}
\vspace{-1mm}
\input{tab/baseline_comp}
\input{tab/ablation}

We compared the CRT and the SAT model\cite{xu2015show}, the ORT\cite{herdade2019image}, and the ABEN\cite{ogura2020alleviating}. Table~\ref{tab:results} shows the quantitative results. We conducted five experimental runs for each method, and the table shows the mean and standard deviation for each metric.

The SAT was used as a baseline because it is one of the most basic models for the image captioning task. We compared the ORT and the CRT to confirm how much the CRB contributed to performance. We also compared the CRT and the ABEN, which did not use the Transformer encoder--decoder architecture. The SAT and the ORT were trained for 20 epochs. We show the result with the best parameter setting on the PFN-PIC dataset\cite{hatori2018interactively} regarding the ORT. The ABEN was trained for 30 epochs. We used 4,000 samples in the training set for the ABEN because when we used all samples,  the training did not satisfy the termination condition.

In this evaluation, we used the standard metrics described below. The primary metric are SPICE and CIDEr-D\cite{vedantam2015cider}, that are standard metrics for image captioning tasks. 
BLEU4\cite{papineni2002bleu} shows the n-gram precision of the generated sentence with the respect to the reference. ROUGE-L\cite{lin2004rouge} is mainly used for summarization and shows the rate of concordance rate with the reference. METEOR\cite{banerjee2005meteor} considers the precision and recall of the generated sentences with respect to unigrams. CIDEr-D is used for image captioning, and it shows the similarity between a generated sentence and a reference sentence. SPICE is another metric for image captioning tasks. 

First, we compare the CRT and the SAT. In Table~\ref{tab:results}, the SPICE and the CIDEr-D were improved by 35.5 and 81.7 points, respectively. The other three metrics scores were also improved. 

Next, we compare the CRT and the ABEN. The table shows that the SPICE and the CIDEr-D were drastically improved by 20.6 and 78.4 points, respectively. This indicates that the Transformer encoder--decoder was an effective approach to the FIG task.

Finally, we compare the CRT and the ORT. The table shows that the SPICE and the CIDEr-D were drastically improved by 17.3 and 67.3 points, respectively. The CRT outperformed the ORT with respect to the other three metrics as well. This indicates that the CRT, using the CRB, successfully handled the referring expressions of a target object and destination.


\subsection{Ablation studies}
We conducted ablation studies on the types of input information. Table~\ref{tab:ablation} quantitatively shows the mean and standard deviation of five experimental runs. We investigated which input features contribute the most to the performance improvement. We considered the seven conditions (a) to (g) shown in Table~\ref{tab:ablation} regarding the combination of the input information $\bm{x}^{<targ>}$, $\bm{x}^{<dest>}$, and $\bm{X}^{<cont>}$. 

Comparing the conditions (a) and (b) with the condition (c), the SPICE  decreased by 6.1 and 3.9 points, respectively. Comparing the condition (d) with the condition (e) and (f),  the SPICE decreased by 1.5 and 9.5 points, respectively. These results indicate that the $\bm{x}^{<targ>}$ contributed the most to performance improvement. 

Comparing the condition (a) with the condition (b), the SPICE decreased by 2.2 points. Comparing the condition (e) with the condition (f), the SPICE decreased by 8.0 points. From this, $\bm{x}^{<dest>}$ contributed to the performance improvement more than $\bm{X}^{<cont>}$. On the other hand, comparing the condition (f) with the condition (g), the SPICE decreased by 1.5 points. From this, $\bm{X}^{<cont>}$ also contributed to the performance improvement.

\begin{figure*}
    \centering
    \includegraphics[clip,height=90mm]{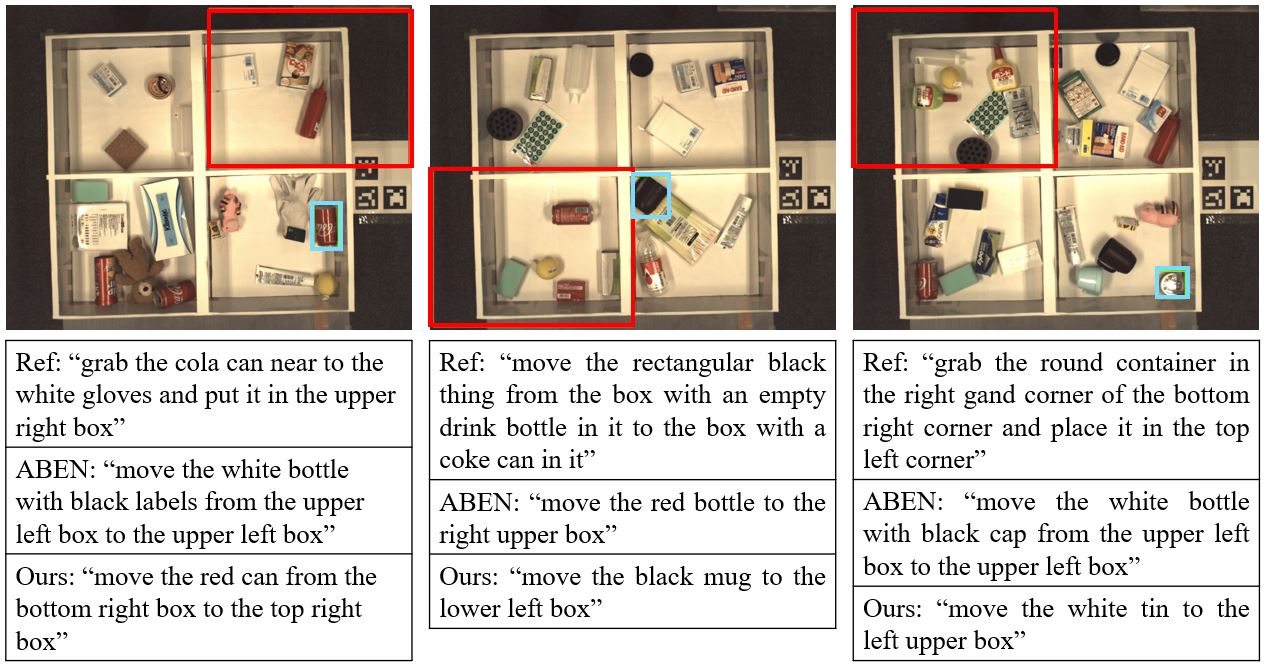}
    \vspace{-2mm}
    \caption{Three samples of qualitative results. Top figures show input images with bounding boxes of target objects (light blue) and destinations (red). Bottom tables show a reference sentence and sentences generated by the ABEN and CRT. }
    \label{fig:result1}
    \vspace{-6mm}
\end{figure*}

\subsection{Qualitative results}
\vspace{-1mm}
Fig.~\ref{fig:result1} shows the qualitative results. 
In the figure, the top panels show the input scenes, and the bottom box shows a reference sentence and the sentences generated by the ABEN and the CRT. The light blue and red bounding boxes represent the target object and the destination, respectively. 

The left-hand sample in the figure shows a sentence that was generated successfully. In the sample, the target object and the destination are ``the cola can in the bottom right box,'' and ``top right,'' respectively. In this sample, there are two Coke (referred to as ``cola'') cans, therefore, an instruction sentence must clearly specify the target object. In the reference sentence, the can was specified as ``the cola can near to the white gloves.'' In the sentence generated by the CRT, the target object was referred to as ``the red can from the bottom right box.'' This indicates that the CRT succeeded in identifying it using a spatial referring expression that was different from reference but still reasonable. On the other hand, in the generated sentence by the ABEN, the target object and the destination were referred to as ``the white bottle with  black labels from the upper left box'' and ``the upper left box,'' respectively. Neither of these expressions is correct.

Similarly, the second sample in the middle column illustrates the successful results obtained by the CRT. In the sample, the target object and the destination is ``black mug,'' and ``bottom left,'' respectively. The reference sentence described the target object as ``the rectangular black thing,'' which was ambiguous. On the other hand, in the generated sentence by the CRT, the target object was referred to as ``the black mug,'' which was valid. In the generated sentence by the ABEN, both the target object and the destination were incorrectly referred to as ``the red bottle'' and ``the right upper box.''

The right-hand sample illustrates a failure case of the CRT. In the sample, the target object and the destination were ``silver can,'' and ``top left,'' respectively. The reference sentence referred to the target object and the destination as ``the round container in the right gand corner of the bottom right corner'' and ``the top left corner,'' respectively. In this sentence, ``gand'' is a misspelling of ``hand.'' Moreover, ``in the right gand corner of the bottom right corner'' is redundant, i.e., ``of the bottom right corner'' is enough. The generated sentence by the baseline referred to the target object as ``the white bottle with black cap from the upper left box.'' This is incorrect because the target object is not ``the white bottle with black cap.'' Furthermore, the target object is not in ``the upper left box.'' The sentence generated by the CRT referred to the target object as ``the white tin.'' It indicates that this is because the target object looks white due to the reflection of the light.

\subsection{Subject experiment}
We conducted a subject experiment in which we compared the reference sentences, sentences generated by the baseline\cite{ogura2020alleviating}, and sentences generated by the CRT. We used the Mean Opinion Score (MOS) as a metric.

Five subjects in their twenties participated in the experiment. Researchers and students specializing in crossmodal language processing were excluded from the subjects to avoid biased evaluation by experts. We asked the subjects to perform the evaluation at their own work speed.

In the experiment, 50 images were randomly selected from the test set. Those images and reference and generated sentences were randomly presented to the subjects. The subjects evaluated the sentences in terms of their intelligibility on a 5-point scale as follows:

\vspace{1mm}
1: Very bad,  2: Bad, 3: Normal, 4: Good, 5: Very good.
\vspace{1mm}

\input{tab/mos}

Table~\ref{tab:mos} shows the mean and standard deviation of the MOS. In Table~\ref{tab:mos}, the MOS of reference sentences was 3.8, which was considered to be the upper bound in this study. The MOS of the CRT was 2.6 and the MOS of the ABEN was 1.2. Statistical significance between the CRT and the ABEN is $ p=8.8 \times 10^{-35}$ $(<0.001)$. It is indicated that the CRT outperformed the ABEN in the subjects experiment as well.

\subsection{Error Analysis}
We analyzed failure cases by the CRT below. We randomly selected 100 samples from the generated sentences for 898 samples of the test set, and confirmed a total of 43 failures in 42 samples.

Table~\ref{tab:erroranalysis} shows the error analysis. We categorized the erroneous generated sentences based on main cause as follows:
\begin{itemize}
    \item{STE (serious target object error):} The generated sentence contains a serious error regarding the color and/or shape of the target; e.g., white gloves were represented as ``white and green box.'' 
    \item{MTE (Minor target object error):} The generated sentence contains a minor error regarding the color or shape of the target; e.g., blue and white striped sandals were represented as ``blue and white tube.''
    \item{DE (Destination error):} A generated sentence contains an incorrect expression about the destination; e.g., the top-right box in the image was incorrectly expressed as ``the left lower box.''
    \item{SE (Spatial referring expression error):} A generated sentence contains an incorrect spatial referring expression to identify the target object and destination; e.g., the destination was expressed as ``to the box with the blue flip flop'' even though there were no blue sandals. 
    \item{O (Others):} This category includes cases where an instruction sentence contains other errors.
\end{itemize}

From Table~\ref{tab:erroranalysis}, the major error of the CRT is STE. Therefore, there is room for improvement in the process of correctly expressing the characteristics of the target object. In particular, some improvement in the image preprocessing (e.g., using EfficientNet\cite{tan2019efficientnet} or Vision Transformer\cite{dosovitskiy2020image} instead of the ResNet-50) may be envisioned to decrease the number of STE.
\input{tab/erroranalysis}

%% file: tab/params.tex
\begin{table}[b]
    \centering
    \caption{Parameter settings of the CRT}
    \vspace{-2mm}
    \normalsize
    \begin{tabular}{|c|l|}
        \hline
         Optimizer & Adam ($\beta_1=0.9, \beta_2=0.999$)  \\ \hline
         Learning rate & $5.0 \times 10^{-4}$ \\ \hline
        Batch size & 15 \\ \hline
        $\#$Epochs & 10 \\ \hline
        $\#$Layers & 6 \\ \hline
        $\#$Attention heads & 8 \\ \hline
    \end{tabular}
    \label{tab:parameters}
    \vspace{-3mm}
\end{table}

%% file: tab/baseline_comp.tex
\begin{table}[tb]

\centering
\caption{Quantitative comparison. The methods have been compared on the PFN-PIC dataset by using the five standard metrics. The mean and standard deviation were calculated on five experimental runs. The best scores are in bold.}

\begin{adjustbox}{width=\linewidth}
\centering
\normalsize
\begin{tabular}{@{}l|ccccc@{}}
\toprule

Method      & BLEU4 & ROUGE-L & METEOR & CIDEr-D & SPICE   \\ \midrule 




SAT~\cite{xu2015show}
& $9.6{\scriptscriptstyle \pm0.5}$    
& $36.1{\scriptscriptstyle \pm0.5}$ 
& $17.2{\scriptscriptstyle \pm0.3}$
& $15.4{\scriptscriptstyle \pm0.7}$
& $9.4{\scriptscriptstyle \pm0.6}$ \\



ABEN~\cite{ogura2020alleviating}
& $\bm{15.2}{\scriptscriptstyle \pm0.8}$    
& $46.8{\scriptscriptstyle \pm1.0}$ 
& $21.2{\scriptscriptstyle \pm0.8}$
& $18.2{\scriptscriptstyle \pm1.8}$
& $23.4{\scriptscriptstyle \pm2.1}$ \\

ORT~\cite{herdade2019image}
& $8.0{\scriptscriptstyle \pm1.2}$    
& $39.4{\scriptscriptstyle \pm0.7}$ 
& $17.3{\scriptscriptstyle \pm0.7}$
& $27.9{\scriptscriptstyle \pm2.8}$
& $26.4{\scriptscriptstyle \pm1.3}$ \\




Ours
& $14.9{\scriptscriptstyle \pm1.1}$    
& $\bm{49.7}{\scriptscriptstyle \pm1.0}$ 
& $\bm{23.1}{\scriptscriptstyle \pm0.7}$
& $\bm{96.6}{\scriptscriptstyle \pm12.0}$
& $\bm{44.0}{\scriptscriptstyle \pm2.3}$ \\

\bottomrule

\end{tabular}
\end{adjustbox}
\label{tab:results}
\end{table}
\normalsize

%% file: tab/ablation.tex
\begin{table*}[bt]
\centering
\caption{Ablation studies on the types of input information. The methods were compared on the PFN-PIC dataset by using the five standard metrics. The best scores are in bold.}
\normalsize
\begin{tabular}{c|c|c|c|ccccc} \toprule
        \multicolumn{4}{c|}{Ablation condition} & BLEU4 &  ROUGE-L & METEOR &  CIDEr-D & SPICE \\ \cline{1-4}
        Condition & $\bm{X}^{<cont>}$ & $\bm{x}^{<dest>}$ & $\bm{x}^{<targ>}$ & & & & & \\ \midrule 
        (a) & \checkmark & & & $7.1{\scriptscriptstyle \pm 1.8}$ & $39.6{\scriptscriptstyle \pm 0.6}$ & $16.5{\scriptscriptstyle \pm 1.0}$ & $27.6{\scriptscriptstyle \pm 2.3}$ & $27.3{\scriptscriptstyle \pm 2.2}$ \\ 
        (b) & & \checkmark & & $12.6{\scriptscriptstyle \pm 4.6}$ & $42.7{\scriptscriptstyle \pm 4.7}$ & $19.7{\scriptscriptstyle \pm 2.5}$ & $34.0{\scriptscriptstyle \pm 5.5}$ & $29.5{\scriptscriptstyle \pm 6.4}$ \\ 
        (c) & & &\checkmark & $9.5{\scriptscriptstyle \pm 0.7}$ & $43.9{\scriptscriptstyle \pm 0.7}$ & $19.3{\scriptscriptstyle \pm 0.5}$ & $74.6{\scriptscriptstyle \pm 4.9}$ & $33.4{\scriptscriptstyle \pm 1.1}$ \\ 
        (d) &\checkmark&\checkmark & &$13.3{\scriptscriptstyle \pm 1.0}$ & $44.0{\scriptscriptstyle \pm 0.5}$ & $20.4{\scriptscriptstyle \pm 0.4}$ & $37.0{\scriptscriptstyle \pm 2.8}$ & $33.0{\scriptscriptstyle \pm 0.8}$ \\ 
        (e) &\checkmark& &\checkmark &$10.3{\scriptscriptstyle \pm 0.6}$ & $44.6{\scriptscriptstyle \pm 1.1}$ & $19.7{\scriptscriptstyle \pm 0.7}$ & $81.8{\scriptscriptstyle \pm 7.0}$ & $34.5{\scriptscriptstyle \pm 2.5}$ \\
        (f) & &\checkmark & \checkmark &$\bm{14.9{\scriptscriptstyle \pm 1.0}}$ & $49.3{\scriptscriptstyle \pm 1.1}$ & $23.0{\scriptscriptstyle \pm 0.5}$ & $92.1{\scriptscriptstyle \pm 6.1}$ & $42.5{\scriptscriptstyle \pm 2.5}$ \\
        (g) & \checkmark &\checkmark & \checkmark &$\bm{14.9{\scriptscriptstyle \pm 1.1}}$ & $\bm{49.7{\scriptscriptstyle \pm 1.0}}$ & $\bm{23.1{\scriptscriptstyle \pm 0.7}}$ & $\bm{96.6{\scriptscriptstyle \pm 12.0}}$ & $\bm{44.0{\scriptscriptstyle \pm 2.3}}$ \\
        \bottomrule
    \end{tabular}
\label{tab:ablation}
\end{table*}

%% file: tab/mos.tex
\begin{table}[b]
    \centering
    \caption{MOS results in subject experiment. The mean and standard deviation are shown.}
    \vspace{-2mm}
    \normalsize
    \begin{tabular}{l|c}
        \toprule
         Method & MOS \\
          \midrule
        Reference sentences (upper bound) &  $3.8\pm1.2$ \\ 
        ABEN\cite{ogura2020alleviating} (baseline) & $1.2\pm0.4$ \\ 
        Ours & $2.6\pm1.5$ \\ 
        \bottomrule

    \end{tabular}
    \label{tab:mos}
    \vspace{-3mm}
\end{table}

%% file: tab/erroranalysis.tex
\begin{table}[t]
    \centering
    \caption{Error analysis on the CRT. Categories of error types.}
    \vspace{-2mm}
    \normalsize
    \begin{tabular}{|l|l|c|}
        \hline
         Error ID & Description & $\#$Errors\\ \hline 
         STE & Serious target object error & 22  \\ \hline 
         MTE & Minor target object error & 12 \\ \hline
         DE & Destination error & 0 \\ \hline
         SE & Spatial referring expression error & 7 \\ \hline
         O & Others & 2 \\ \hline
         Total & - & 43 \\ \hline
    \end{tabular}
    \label{tab:erroranalysis}
    \vspace{-6mm}
\end{table}

%% file: section7.tex
\vspace{-1mm}
\section{Conclusions}
\vspace{-1mm}
Most data-driven approaches for crossmodal language understanding require large-scale datasets.  
However, building such a dataset is time-consuming and costly.
We hence proposed the Case Relation Transformer (CRT), a crossmodal language generation model that can generate an instruction sentence that includes  referring expressions with a target object and destination.

Our contributions are as follows:
\begin{itemize}
 \item In the CRT, the Case Relation Block was introduced to handle the relationships between the target object, destination, and context information.
 \item  The CRT outperformed the baseline models on five standard metrics in the fetching instruction generation task.
 \item The results of the subject experiment using the MOS also shows the CRT could generate statistically significantly better sentences than the ABEN\cite{ogura2020alleviating}.

\end{itemize}

In future research, we will train a crossmodal language understanding model using the dataset augmented by the proposed method.